\documentclass{article}
\usepackage{iclr2027_conference,times}

\usepackage{amsmath,amsfonts,bm}

\def\eqref#1{equation~\ref{#1}}
\def\1{\bm{1}}

\DeclareMathAlphabet{\mathsfit}{\encodingdefault}{\sfdefault}{m}{sl}
\SetMathAlphabet{\mathsfit}{bold}{\encodingdefault}{\sfdefault}{bx}{n}

\usepackage{hyperref}
\usepackage{url}
\usepackage{booktabs}
\usepackage{amsmath}
\usepackage{amssymb}
\usepackage{algorithm}
\usepackage{algorithmic}
\usepackage{multirow}
\usepackage{xcolor}
\usepackage{colortbl}
\usepackage{graphicx}
\usepackage{subcaption}
\usepackage{tcolorbox}
\tcbuselibrary{skins,breakable}

\newtcolorbox{skillbox}[1]{%
  enhanced, breakable,
  colback=white,
  colframe=black!70,
  colbacktitle=black!75,
  coltitle=white,
  fonttitle=\bfseries\small,
  title=#1,
  boxrule=0.4pt,
  arc=2.5pt,
  toprule=0pt,
  leftrule=0.4pt,
  rightrule=0.4pt,
  bottomrule=0.4pt,
  titlerule=0pt,
  left=6pt, right=6pt, top=5pt, bottom=5pt,
  before skip=8pt, after skip=10pt,
}

\definecolor{bestpurple}{HTML}{E0D2F1}
\definecolor{secondpurple}{HTML}{F6F1FB}
\newcommand{\bestcell}[1]{\cellcolor{bestpurple}\textbf{#1}}
\newcommand{\secondcell}[1]{\cellcolor{secondpurple}#1}

\newcommand{\method}{\textsc{Selm}}
\newcommand{\lib}{\mathcal{L}}
\newcommand{\skill}{s}
\newcommand{\skills}{\mathcal{S}}
\newcommand{\graveyard}{\mathbf{G}}
\newcommand{\crud}{\mathcal{A}}

\title{\textsc{Selm}: Self-Evolving Library Management for \\
Multi-Skill Agent Optimization}

\author{%
Mengxi Hu$^{1,2}$,
Yuxuan Hu$^{1,2}$,
Jinbo Su$^{1,2}$,
Ke Wang$^{1,2}$,
Honghao Shi$^{5}$,
Wenli Wu$^{5}$,\\
Cuiping Li$^{1,4}$,
Hong Chen$^{1,3}$,
Jing Zhang$^{*1,2}$\\[0.5em]
$^{1}$School of Information, Renmin University of China, Beijing, China\\
$^{2}$Key Laboratory of Data Engineering and Knowledge Engineering, Beijing, China\\
$^{3}$Engineering Research Center of Database and Business Intelligence, Beijing, China\\
$^{4}$Beijing Key Laboratory of Intelligent Database Systems and Applications, Beijing, China\\
$^{5}$Beijing Academy of Artificial Intelligence, Beijing, China\\[0.3em]
\texttt{\{humengxi0206,zhang-jing\}@ruc.edu.cn}%
}

\iclrfinalcopy
\begin{document}

\maketitle

\begin{abstract}
%Large language model (LLM) agents increasingly use textual skills as a lightweight alternative to model finetuning for adapting to diverse tasks. Recent methods optimize skills from execution trajectories, but primarily focus on evolving a single skill or accumulating multiple skills without systematically managing their retrieval, updates, and lifecycle. This limits their ability to isolate conflicting experience, select useful skills, and remove ineffective ones as the skill collection grows.
%We present \method{} (Self-Evolving Library Management), a framework that formulates skill evolution as library-level management. \method{} maintains a structured skill library and jointly manages which skills to retrieve, which skills to update from execution experience, and which skills to retain or retire. Specifically, it combines utility-aware retrieval, Map-Reduce skill update, and lifecycle controls to continuously evolve a compact and specialized skill library without updating model parameters.We evaluate \method{} on four agent benchmarks covering diverse tasks and execution environments, and \method{} consistently outperforms in most settings.%no-skill agents and single-document skill optimization, while remaining competitive with recent multi-skill methods, demonstrating the effectiveness of library-level skill management for continual agent adaptation.

Large language model (LLM) agents increasingly use  skills as a lightweight way to acquire task-specific knowledge. 
Recent work enables skills to evolve from task trajectories. However, these methods often lack systematic management of evolved skills, making it difficult for agents to reuse and accumulate skills across domains, thus limiting performance on subsequent tasks.
To reliably accumulate, reuse, and maintain useful skills across tasks, we present \method{} (Self-Evolving Library Management), a framework that formulates skill evolution as library-level management. \method{} jointly manages skill retrieval, updating, and lifecycle by maintaining skill confidence and utility signals. 
It introduces utility-aware retrieval to select useful skills, a Map-Reduce framework to generate targeted updates from task-specific experience, and lifecycle controls to retain or remove skills based on their effectiveness. 
These mechanisms enable the skill library to continuously accumulate useful experience while preserving and pruning skills across tasks, improving agent performance on complex tasks across diverse domains. We evaluate \method{} on four agent benchmarks covering diverse tasks and execution environments, and \method{} consistently outperforms in most settings.

\end{abstract}

% ============================================================
\section{Introduction}
\label{sec:intro}
% ============================================================

Large language model (LLM) agents are increasingly used to solve complex tasks across different domains. To improve agents' behaviors on specific tasks, recent work introduced skills~\citep{anthropic2025agentskills}, which are pieces of text that can be plugged into the agent context to provide task-specific knowledge, procedures, or experience. This gives agents a simple way to gain specialized knowledge without additional model training.

Recent work has explored how skills can evolve as the agent runs. Early skills were mostly static: they were written by experts or saved from successful experiences and then kept unchanged~\citep{zhao2024expel,anthropic2025agentskills}. More recent methods allow agents to update skills from their execution trajectories~\citep{yang2026skillopt,alzubi2026evoskill,liu2026skillforge,zhang2026skillevolver,zheng2025skillweaver,ni2026trace2skilldistilltrajectorylocallessons,ma2026skillgen,zhou2026mementoskills,zhang2026coevoskills}. After running tasks, the agent can examine which skill worked or failed, then edit existing skills or create new ones. In this way, skills can change with the agent's experience and adapt to new tasks and environments.  

\begin{figure*}[t]
\centering
\includegraphics[width=\textwidth]{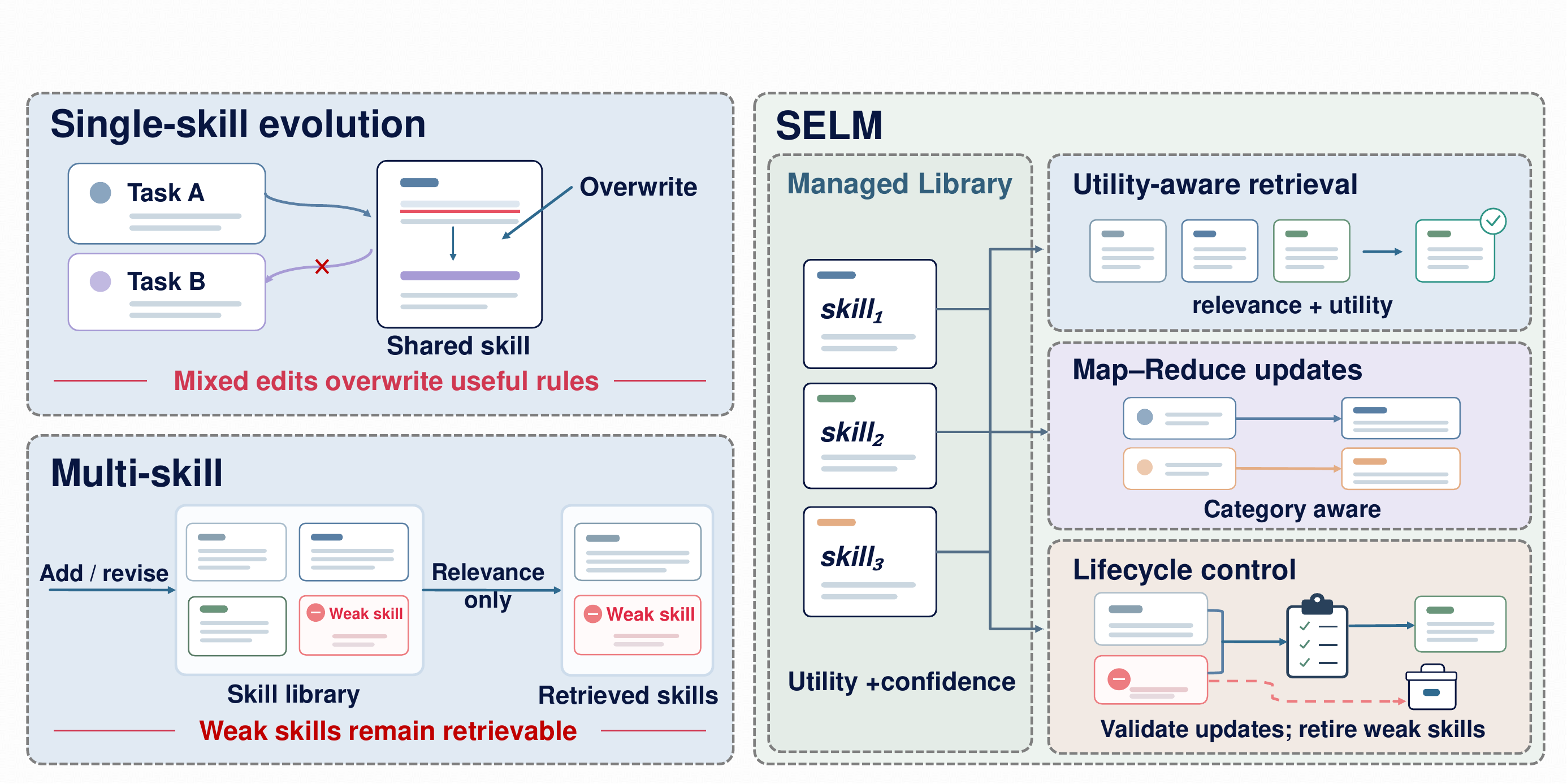}
\caption{Prior skill-evolution methods can overwrite useful procedures or keep weak skills retrievable, causing task failures when experience from different tasks conflicts. \method{} instead uses utility-aware retrieval, category-wise Map–Reduce updates, and lifecycle validation to preserve useful skills and retire ineffective ones, enabling more reliable task completion.}
\label{fig:intro}
\end{figure*}

However, although these evolution methods propose many strategies to evolve skills ~\citep{alzubi2026evoskill,zhang2026skillevolver,zheng2025skillweaver,zhou2026mementoskills,zhang2026coevoskills}, 
%few of them manage the skills already in hand throughout the course of evolution
few of them focus on how to manage them.
A skill revision that best fits the current round can be harmful to later processes. 
For instance, a skill that is unused or unhelpful for the present task may be overwritten or discarded; once that change is committed, a later task can never recover the deleted skill. 
Meanwhile, failed or redundant skills that have remained in the system may still be loaded, giving the agent wrong instructions. 
The core problem is that existing methods always treat skill updates as separate, one-round edits, rather than managing skills as a dynamic, long-term library. %for the whole evolving process. 
Thus, a skill management library is needed to decide not only how a skill should change, but also which skills should be retrieved, which skills should be updated, and which skills should remain in the library.

To address these limitations, we propose \method{} (Self-Evolving Library Management), a framework that manages skills as a library and jointly controls their retrieval, updates and lifecycle. By tracking and leveraging skill confidence and utility, \method{} enables the library to evolve with task experience and improves the performance of the agent across complex tasks in different domains.

Specifically, \method{} introduces a \textbf{skill library} with explicit mechanisms for managing skill states. 
To dynamically evaluate skill usefulness, each skill is associated with a confidence score and a utility score. 
Confidence reflects how strongly the skill is supported by update operations, while utility reflects how useful it has been across tasks. %They are used to decide which skills stay active, which are removed, and which failed skills should not be added again. 
These mechanisms provide a clear basis for deciding which skills should stay active, which skills should be removed, and which failed skills should not be added again. The library can thus keep a skill that is unused on the current task but still useful later, and drop one that keeps failing, instead of committing every local edit, enabling reliable, long-term skill management.

Second, \method{} designs a \textbf{Map-Reduce framework} for skill updates. Applying every trajectory to all the skills will make experience from different types of task overwrites each other. \method{} therefore groups execution trajectories by task category(Map) and analyzes each group separately, identifying which skills are affected and what changes are needed. It then combines these analyses into operations(Reduce), such as adding new skills, editing existing ones, and deleting those that no longer help. All these operations apply only to the relevant skills and unrelated skills are left unchanged, so a change that helps the current category won't erase skills required later, which enhances skill effectiveness.

%Third, \method{} proposes \textbf{utility-aware skill retrieval}. Existing methods mainly retrieve by semantic relevance~\citep{yang2026autoskill}, so a skill can look related to the current task even though it has performed poorly before, and the agent may still follow it. \method{} first selects skills from the same category, then ranks them by both task relevance and historical utility. This way of ranking helps distinguish skills that are not only related to the current task but also have a history of being useful.

Third, \method{} proposes \textbf{utility-aware skill retrieval}. Existing methods mainly retrieve skills based on their semantic relevance to the current task~\citep{yang2026autoskill},
%. However, semantic relevance only indicates whether a skill looks related to the current task, and it does not indicate 
which overlooks whether the skill has been useful in previous tasks. 
%\method{} therefore uses both current task relevance and historical utility for skill retrieval. 
\method{} leverages historical utility to enhance skill retrieval.
It first selects skills from the same task category and then ranks them by their relevance to the current task and their past usefulness. %This helps the agent distinguish skills that are both relevant to the current task and shown useful through previous execution experience, thus improving agent's performance on current task.
This helps the agent distinguish skills that are relevant to task on hand and have proven useful through previous execution experience, thus enabling more effective skill selection.

Finally, we evaluate \method{} on four agent benchmarks covering diverse execution environments and areas. \method{} achieves state-of-the-art results among skill-evolution methods, outperforming both single-document optimization and prior multi-skill baselines without updating model parameters.

Our contributions are (1)~introducing a \textbf{skill library} that tracks skill confidence and utility, managing skill validation, retention, and removal, enabling reliable skill management throughout the evolution process; (2)~designing a \textbf{Map-Reduce framework} that analyzes trajectories by task category and generates targeted operations, allowing the agent to update relevant skills based on task-specific experience; and (3)~developing a \textbf{utility-aware retrieval} method that ranks skills using both task relevance and historical utility, enabling the agent to select skills that are both relevant and useful. Across our experiments, \method{} outperforms existing skill optimization methods, demonstrating the effectiveness of the proposed framework.

% ============================================================
\section{Related Work}
\label{sec:related}
% ============================================================
\paragraph{Skill construction and evolution.}
Recent work has increasingly explored constructing and evolving textual skills from agent experience, moving beyond manually specified or static skill representations~\citep{zhao2024expel,anthropic2025agentskills}.
More broadly, approaches such as TextGrad~\citep{yuksekgonul2025textgrad} and GEPA~\citep{agrawal2026gepa} optimize textual instructions through feedback without updating model parameters, while Dynamic Cheatsheet~\citep{suzgun2026dynamiccheatsheet} and ACE~\citep{zhang2025ace} iteratively refine textual context from agent experience.
Building on this paradigm, SkillOpt and SkillGrad optimize a shared skill document or skill package by repeatedly editing textual knowledge based on execution feedback~\citep{yang2026skillopt,wang2026skillgrad}.
Other approaches distill execution trajectories into standalone lessons, strategic principles, or verified skills, thereby converting individual experiences into reusable knowledge~\citep{zhao2024expel,wu2026evolver,ni2026trace2skilldistilltrajectorylocallessons,ma2026skillgen,qiu2026autorefine}.
Rather than maintaining a single shared artifact, SkillWeaver~\citep{zheng2025skillweaver}, EvoSkill~\citep{alzubi2026evoskill}, SkillEvolver~\citep{zhang2026skillevolver}, Memento-Skills~\citep{zhou2026mementoskills}, and CoEvoSkills~\citep{zhang2026coevoskills} construct collections of specialized skills to capture different task types or capabilities.
Complementary approaches construct skills from external domain resources or learn skill-related policies through interaction and co-training~\citep{yang2026autoskill,liu2026skillforge,xia2026skillrl,jiang2026xskill}.
Despite these advances, existing methods primarily focus on how skills are created or improved, treating skill evolution as a sequence of local construction or editing operations. This becomes limiting when multiple skills coexist: updates to a shared artifact can introduce conflicting changes, while independently accumulated skills provide limited mechanisms for determining which skills should be updated.

\paragraph{Skill library management.}
Early work such as Voyager~\citep{wang2023voyager} and ExpeL~\citep{zhao2024expel} retrieves executable or experiential knowledge according to the relevance of the current task.
Subsequent systems make skill selection more behavior-aware and introduce mechanisms for validating candidate skills or deciding whether updates should be merged, accepted, or discarded~\citep{zhou2026mementoskills,yang2026skillopt,yang2026autoskill,alzubi2026evoskill,zhang2026coevoskills,qiu2026autorefine}.
More recent work explicitly considers the maintenance of the skill library itself. SkillOps studies how an existing skill collection can be periodically reorganized to address issues that emerge as the library grows~\citep{pu2026skillops}, while MUSE-Autoskill organizes skill creation, catalog retrieval, per-skill memory, and test-driven refinement into successive lifecycle stages~\citep{lin2026museautoskill}.
Lifelong evaluations further suggest that usage frequency alone is insufficient to characterize the utility of a skill, motivating more explicit utility-aware management~\citep{zhang2026skillflow}.
However, these approaches generally address retrieval, validation, update selection, or lifecycle maintenance as separate mechanisms, rather than jointly optimizing them as coupled decisions over a continually evolving skill library.
%\method{} instead treats the library as the central optimization object, jointly considering utility-aware retrieval, trajectory-driven updates, and lifecycle management to coordinate which skills are used, improved, retained, or retired over time.

% ============================================================
\section{Preliminary}
\label{sec:formulation}
% ============================================================

This section introduces the basic concepts of skills, skill retrieval, and skill evolution.

\paragraph{Skill.}
The $k$-th skill is represented as
\begin{equation}
\skill_k = (n, d, c, t, \gamma),
\label{eq:skill}
\end{equation}
where \(n(s)\), \(d(s)\), \(c(s)\), \(t(s)\), and \(\gamma(s)\) denote the skill name, description, instruction content, trigger conditions, and task category, respectively.

As the number of skills grows, they form a skill library, which is a collection of reusable skills that the agent can draw on for different tasks. Formally,
\begin{equation}
%\skills = \{\skill_1, \dots, \skill_{|\skills|}\}.
\lib = \{\skill_1, \dots, \skill_{|\lib|}\}.
\label{eq:library}
\end{equation}

\paragraph{Skill retrieval.}
The agent does not need to use the entire library for every task. Given a task $x$, a retrieval mechanism selects a subset of skills from the library to provide relevant instructions for the task. We denote the retrieved skill set as
\begin{equation}
\skills(x) = \operatorname{Retrieve}(\lib, x),
\qquad
\skills(x) \subseteq \lib.
\label{eq:retrieval}
\end{equation}
The selected skills are then provided to the agent as part of its context before task execution.

\paragraph{Evolving skills.}
With frozen target model $M$ and current library, the agent runs on task $x$, and the resulting experience is written back so that the same library can improve over iterations. $r(\lib, x)$ represents the score of library $\lib$ on task $x$. Given training, selection, and test splits $D_{\mathrm{tr}}$, $D_{\mathrm{sel}}$, and $D_{\mathrm{test}}$, evolution proposes libraries from $D_{\mathrm{tr}}$ and keeps the one that scores best on $D_{\mathrm{sel}}$. Formally,
\begin{equation}
\lib^* = \arg\max_{\lib \in \mathcal{C}(D_{\mathrm{tr}})} \frac{1}{|D_{\mathrm{sel}}|} \sum_{x \in D_{\mathrm{sel}}} r(\lib, x), 
\qquad
r(\lib, x) \in [0,1]
\label{eq:selection}
\end{equation}
where $\mathcal{C}(D_{\mathrm{tr}})$ is the set of libraries obtained by evolving on the training split. 
%We report $D_{\mathrm{test}}$ performance of $\lib^*$. 
The score $r(\lib, x)$ and the operators that map trajectories to library edits are instantiated in \S\ref{sec:method}.

\paragraph{CRUD operations.}
CRUD is an operational concept in databases for managing persistent records. It stands for Create, Read, Update, and Delete. We migrate this concept to skill management. In this paper, the library operations are $\mathrm{ADD}$, $\mathrm{EDIT}$, $\mathrm{UPVOTE}$, $\mathrm{DOWNVOTE}$, and $\mathrm{DELETE}$: $\mathrm{ADD}$ creates a new skill, $\mathrm{EDIT}$ revises an existing skill, $\mathrm{UPVOTE}$ and $\mathrm{DOWNVOTE}$ adjust its confidence, and $\mathrm{DELETE}$ removes it from the active library.

% ============================================================

% ============================================================
\section{Method}
\label{sec:method}
% ============================================================
\method{} freezes the target model $M$ and optimizes only the skill library. 
At each iteration, the agent retrieves a small set of skills from the current library, uses them to solve a batch of training tasks, and converts the resulting trajectories into library updates. 
The updated library is then evaluated on the selection set and retained only when it passes the validation gate. 
This iterative process jointly improves skill retrieval, skill updates, and library quality, with the overall procedure summarized in Algorithm~\ref{alg:main} (Appendix~\ref{app:algorithm}).

\begin{figure}[t]
\centering
\includegraphics[width=\linewidth]{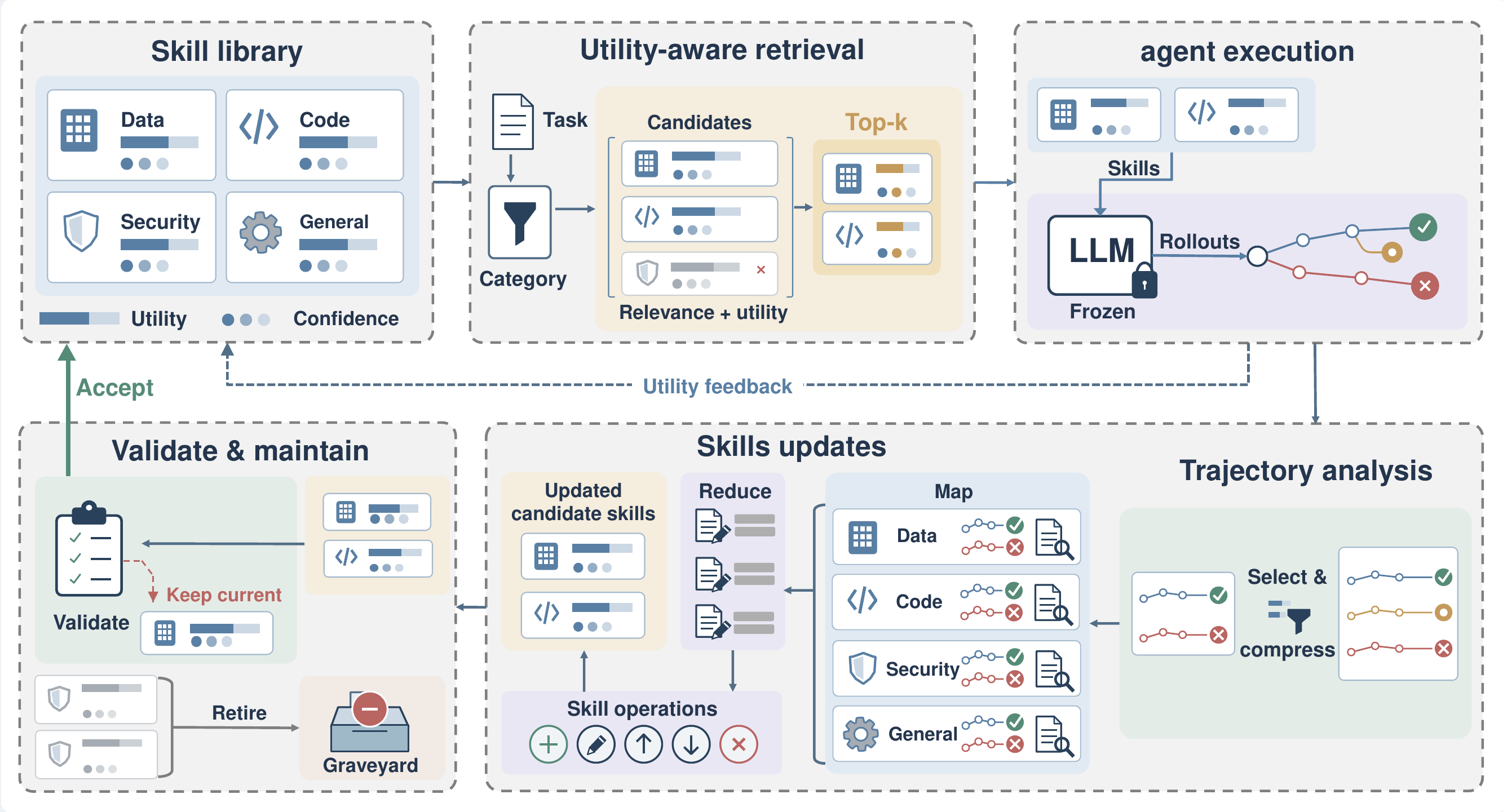}
\caption{Overview of \method{}. The frozen agent retrieves skills by relevance and utility, executes tasks, and turns trajectories into category-wise Map--Reduce updates. Candidate skills are validated on a held-out set; accepted skills remain in the library, and low-utility skills are retired to a graveyard that blocks their names from returning.}
\label{fig:overview}
\end{figure}

For each skill, \method{} manages it through two pieces of information: confidence score and utility score. 
The confidence score $v(\skill)$ reflects how strongly a skill should be retained based on update operations, while the utility score $u(\skill)$ records how useful the skill has been across tasks. 
These two scores support different stages of the library evolution process: confidence is updated by UPVOTE/DOWNVOTE operations and used for lifecycle control, while utility is used to guide skill retrieval and updated from execution outcomes. 
Their corresponding update mechanisms are described in \S\ref{sec:retrieval} and \S\ref{sec:evolution}, respectively.

\subsection{Utility-Aware Skill Retrieval}
\label{sec:retrieval}

\method{} retrieves skills by jointly considering task category, task relevance, and historical utility. For each task $x_i$, let $\gamma_i$ denote its task category. \method{} first constructs a candidate set from the active skill set $\skills_t$, retaining skills whose categories match $\gamma_i$ or are marked as ``general''. It then ranks the candidates using a utility-aware retrieval score and selects the top-$k$ skills for injection. Formally,
\begin{equation}
R(\skill,x_i)
=
(1-w_u)R_{\mathrm{rel}}(\skill,x_i)
+
w_u u(\skill),
\label{eq:score}
\end{equation}
where $R_{\mathrm{rel}}(\skill,x_i)$ measures the relevance between skill $\skill$ and task $x_i$ using semantic and keyword-based matching, and $u(\skill)$ is the historical utility of the skill. 
The coefficient $w_u$ controls the contribution of historical utility.

The top-$k$ skills are selected for task $x_i$:
\begin{equation}
\skills_t(x_i)
=
\operatorname{top}\text{-}k
\left(
\left\{
\skill\in\skills_t
\mid
\gamma(\skill)\in\{\gamma_i,\text{``general''}\}
\right\},
R(\skill,x_i)
\right).
\label{eq:retrieve}
\end{equation}
The selected skills $\skills_t(x_i)$ are injected into the agent context before execution. 
This retrieval strategy favors skills that are both relevant to the current task and useful based on previous execution results.

\textbf{Utility Feedback.}
The utility score measures how useful a skill has been on the tasks that used it. 
Let $\mathcal{T}(\skill)$ denote the set of tasks whose retrieved skill set contains $\skill$. 
The average reward of $\skill$ in the current update is
\begin{equation}
\bar{r}_{\skill}
=
\frac{1}{|\mathcal{T}(\skill)|}
\sum_{x\in\mathcal{T}(\skill)}
r(\lib_t,x).
\label{eq:skill_reward}
\end{equation}
We update the utility using an exponential moving average (EMA)
\begin{equation}
u(\skill)
\leftarrow
(1-\alpha)u(\skill)
+
\alpha\bar{r}_{\skill},
\label{eq:ema}
\end{equation}
where $\alpha$ controls the contribution of new observations. The same utility score is then used by the retrieval mechanism in the next iteration.
Thus, utility provides a feedback signal across successive retrieval rounds: execution outcomes update the utility of the skills that were used, and the resulting utility is subsequently incorporated into retrieval, reducing the selection of ineffective skills and improving agent performance across tasks.

\subsection{Skill Library Evolution}
\label{sec:evolution}

After retrieving skills for each task, \method{} runs the frozen target model and uses the resulting trajectories to update the skill library. 
The update process consists of four steps: rollout, trajectory analysis, CRUD synthesis, and candidate construction.

\textbf{Rollout and Reward.}
For each training task $x_i$, the selected skills $\skills_t(x_i)$ are injected into the agent context. 
Given the frozen target model $M$ and execution harness $h$, the rollout produces a trajectory and task scores:
\begin{equation}
\left(
\tau_i,
r_{\mathrm{hard},i},
r_{\mathrm{soft},i}
\right)
=
h\left(
M,x_i,\skills_t(x_i)
\right),
\label{eq:single_exec}
\end{equation}
where
$r_{\mathrm{hard},i}\in\{0,1\}$
indicates whether the task $x_i$ is fully solved, and
$r_{\mathrm{soft},i}\in[0,1]$
measures partial progress in two ways:
When the task $x_i$ is judged by multiple checks,
$r_{\mathrm{soft},i}=n_{\mathrm{pass},i}/n_{\mathrm{total},i}$, and
$r_{\mathrm{hard},i}=1$ only when every check passes.
When a task is judged by comparing a prediction with a reference answer,
$r_{\mathrm{soft},i}$ is their token-level overlap and
$r_{\mathrm{hard},i}$ is exact match.

The reward used for library optimization is obtained from these task scores:
\begin{equation}
r_i
=
r(\lib_t,x_i)
=
\mathcal{G}
\left(
r_{\mathrm{hard},i},
r_{\mathrm{soft},i}
\right),
\label{eq:gated_score}
\end{equation}
where
\begin{equation}
\mathcal{G}(r_{\mathrm{hard}},r_{\mathrm{soft}})
=
\begin{cases}
r_{\mathrm{hard}},
& \text{if metric}=\text{hard},\\
r_{\mathrm{soft}},
& \text{if metric}=\text{soft},\\
(1-w)r_{\mathrm{hard}}
+w r_{\mathrm{soft}}, w\in [0,1],
& \text{if metric}=\text{mixed}.
\end{cases}
\label{eq:gate}
\end{equation}
Each rollout therefore yields two signals for library evolution. The trajectory $\tau_i$ is the execution evidence, including the agent's actions, tool calls, intermediate outputs, and errors. The scalar reward $r_i$ shows how much of the task was solved. Later updates read $\tau_i$ to decide what to change, and use $r_i$ to decide which trajectories are worth changing from.

\textbf{Trajectory Selection.} 
\method{} prioritizes trajectories whose observed rewards differ substantially from the historical utility of the retrieved skills. 
For task $x_i$, we compute the average utility of its retrieved skills:
\begin{equation}
\bar{u}(\skills_t(x_i))
=
\frac{1}{|\skills_t(x_i)|}
\sum_{\skill\in\skills_t(x_i)}
u(\skill),
\label{eq:avg_util}
\end{equation}
and define the diagnostic priority as
\begin{equation}
p_i
=
\left|
r_i-\bar{u}(\skills_t(x_i))
\right|.
\label{eq:priority}
\end{equation}
\method{} selects the highest-priority trajectories for further analysis. 
Trajectories with large $p_i$ provide stronger evidence that the retrieved skills may need to be updated. $\bar{u}(\skills_t(x_i))$ is the library's current belief about those skills, and a large gap means the observed reward may contradict that belief. A much higher $r_i$ shows a successful pattern that the stored utility under-credits, and a much lower $r_i$ shows that the retrieved guidance failed on this task. A small gap means the rollout agrees with the existing record, so it adds little new information about what to change. 

% 加不加?
%The selected trajectories are then compressed according to their outcomes. Failed trajectories retain error-related steps and their local context, while successful trajectories are reduced to their essential beginning and ending steps. A reward threshold is also used to distinguish complete failures from partial successes, allowing trajectories with positive partial credit to contribute useful success patterns.

\textbf{Map---Category-wise Analysis.}
The selected trajectories are partitioned by task category and analyzed independently. 
For each category, a \texttt{TrajectoryAnalyst} identifies successful patterns, failure patterns, and new insights from the corresponding trajectories. 
The analysis prevents experience from unrelated task categories from being directly mixed during skill updates.

\textbf{Reduce---Skill update.}
Given the category-wise trajectory analyses and the current skill library, a \texttt{SkillSynthesizer} generates a bounded set of CRUD operations. 
The synthesizer considers the current skill content, confidence, utility, and category-specific execution feedback when deciding how to update the library.

Given a set of trajectories, \method{} updates the library through the CRUD operations introduced in \S\ref{sec:formulation}:
\begin{equation}
\crud =
\{\mathrm{ADD},\mathrm{EDIT},\mathrm{UPVOTE},\mathrm{DOWNVOTE},\mathrm{DELETE}\}.
\label{eq:crud_ops}
\end{equation}

For an existing skill $\skill$, the confidence score is updated as
\begin{equation}
v(\skill) \leftarrow
\begin{cases}
v(\skill)+1, & \text{for } \mathrm{UPVOTE},\\
v(\skill)-1, & \text{for } \mathrm{DOWNVOTE}.
\end{cases}
\label{eq:confidence}
\end{equation}

The generated operations are applied to the relevant skills:
\begin{equation}
\lib'_t
=
\operatorname{Apply}
\left(
\lib_t,\mathrm{ops}_t
\right),
\qquad
\mathrm{ops}_t\subseteq\crud.
\label{eq:apply}
\end{equation}
%The operations can add new skills, edit existing skills, reinforce useful skills, weaken misleading skills, or delete skills that should no longer remain active.

When a category exhibits recurring failure patterns, the synthesizer can also create new skills that describe strategies for avoiding these failures. 
The resulting candidate library $\lib'_t$ is then passed to the validation stage.

\textbf{Multi-timescale Updates.} Following prior work on iterative skill evolution~\citep{yang2026skillopt}, \method{} uses both fast and slow updates. Fast updates are performed after each training batch to incorporate recent trajectory feedback, while slow updates are performed less frequently across epochs to consolidate changes that persist over time. We also allow slow updates to produce meta skills, which summarize persistent cross-skill patterns identified across an epoch. This allows the library to respond to recent task feedback while reducing unnecessary updates caused by short-term fluctuations.

\subsection{Skill Lifecycle Management}
\label{sec:lifecycle}

\method{} treats skill maintenance as an explicit lifecycle process, using validation, utility tracking, and retirement mechanisms to prevent ineffective skills from accumulating or re-entering the library. Throughout optimization, the active set $\skills_t$ of $\lib_t$ and the graveyard $\graveyard_t$ obey three structural invariants:
\begin{align}
|\skills_t| &\leq M_{\max} && \text{(capacity)}, \label{eq:capacity} \\
\forall \skill_i, \skill_j \in \skills_t,\ i \neq j &\implies n(\skill_i) \neq n(\skill_j) && \text{(uniqueness)}, \label{eq:unique} \\
\forall \skill \in \skills_t,\ n(\skill) &\notin \{n(g) \mid g \in \graveyard_t\} && \text{(graveyard blocking)}. \label{eq:grave}
\end{align}
Capacity bounds the store; uniqueness forbids duplicate names in $\skills_t$; graveyard blocking forbids reintroducing a retired name.

\textbf{Validation.} A candidate update is not directly added to the active library. 
\method{} first evaluates the candidate library $\lib'_t$ on the selection set $D_{\mathrm{sel}}$ using the frozen target model and the same execution harness. 
Its validation score is
\begin{equation}
V(\lib'_t)
=
\frac{1}{|D_{\mathrm{sel}}|}
\sum_{x\in D_{\mathrm{sel}}}
r(\lib'_t,x).
\label{eq:validation}
\end{equation}
The candidate library is retained only when it improves the validation performance over the current library. 
This validation gate prevents updates that appear reasonable from the trajectory analysis but reduce the actual task performance from being incorporated into the library.

\textbf{Pruning and Graveyard.} \method{} sets a threshold $\tau_u$ for utility score. For skill $s$, if $u(s)< \tau_u$ or $v(\skill)<0$, it will be retired from the active set $\skills_t$ and moved to the graveyard $\graveyard_t$. Retired skills are excluded from future retrieval and cannot simply be reintroduced under the same identity. This prevents ineffective skills from repeatedly entering and leaving the library and keeps the active library compact.

% ============================================================
\section{Experiments}
\label{sec:experiments}
% ============================================================

We evaluate \method{} as a library-level optimizer for frozen agents.
The experiments answer three questions.
(i) Does a managed skill library improve over no skill, static skills, prompt evolution, single-document skill optimization, and recent multi-skill methods?
(ii) Does that comparison hold across different models?
(iii) Which library-management controls are required for the gains?

\subsection{Experimental Setup}

\textbf{Baselines.} We compare against \textbf{No skill}; \textbf{Human skill} (an expert-written document per benchmark); \textbf{LLM skill} (a one-shot skill from Claude-Opus-4.6 that is never updated); prompt-evolution \textbf{GEPA}~\citep{agrawal2026gepa}; skill-construction methods \textbf{SkillEvolver}~\citep{zhang2026skillevolver}, \textbf{Memento-Skills}~\citep{zhou2026mementoskills}, and \textbf{Trace2Skill}~\citep{ni2026trace2skilldistilltrajectorylocallessons}; single-document \textbf{SkillOpt}~\citep{yang2026skillopt}; and, under Claude Code, folder-based \textbf{EvoSkill}~\citep{alzubi2026evoskill}. EvoSkill is omitted from the no-harness block because it assumes a \texttt{SKILL.md} package layout. Qwen3-30B-A3B is evaluated only without a harness. 

\textbf{Benchmarks.} We use four datasets: SkillsBench~\citep{li2026skillsbench}, SpreadsheetBench~\citep{ma2024spreadsheetbench}, OfficeQA~\citep{opsahlong2026officeqaproenterprisebenchmark}, and LiveMath~\citep{he2026livemathematicianbench}. Detailed descriptions are in Appendix~\ref{app:benchmarks}.

\textbf{Target models.} We use Claude-Opus-4.6~\citep{anthropic2026claudeopus46}, DeepSeek-V4-Pro~\citep{deepseekai2026deepseekv4}, and Qwen3-30B-A3B~\citep{qwen2025qwen3} as frozen target models. Claude and DeepSeek are evaluated both without a harness and inside Claude Code~\citep{anthropic2025claudecode}; Qwen3-30B-A3B is evaluated only without a harness. No-harness runs prepend retrieved skills to the system prompt. Claude Code loads skills through the native agent workspace. All methods in a given cell use the same target model, harness, and scorer.

\textbf{Hyperparameters.} Hyperparameters are reported in Appendix~\ref{app:hyperparams}.

\subsection{Main Results}

\begin{table}[t]
\vspace{-1.6em}
\caption{Main results on SkillsBench, OfficeQA, SpreadsheetBench, and LiveMath. SkillsBench reports soft/hard accuracy (\%); other rows are accuracy (\%). The best cell in each row within a model block is bold and shaded purple; the second-best cell is shaded a lighter purple. Note: Qwen3-30B-A3B failed all the tasks on SkillsBench, so the Hard Acc is not included.
}
\label{tab:main_results}
\centering
\vspace{-0.9em}
\scriptsize
\renewcommand{\arraystretch}{0.76}
\setlength{\tabcolsep}{2.2pt}
\setlength{\aboverulesep}{0.28ex}
\setlength{\belowrulesep}{0.32ex}
\textbf{No harness}\\[-0.08em]
\resizebox{\linewidth}{!}{%
\begin{tabular}{@{}llccccccccc@{}}
\toprule
\multicolumn{2}{@{}l}{\textbf{Benchmark}} & \textbf{No skill} & \textbf{Human skill} & \textbf{LLM skill} & \textbf{GEPA} & \textbf{SkillEvolver} & \textbf{Memento-Skills} & \textbf{SkillOpt} & \textbf{Trace2Skill} & \textbf{\method{}} \\
\midrule
\multicolumn{11}{@{}l}{\textit{Claude-Opus-4.6}} \\
\multirow{2}{*}{SkillsBench} & Soft Acc & 40.4 & 45.3 & 44.8 & 56.4 & 59.3 & \secondcell{66.1} & 62.7 & 52.4 & \bestcell{69.8} \\
 & Hard Acc & 31.5 & 36.3 & 34.5 & 41.8 & 43.6 & \secondcell{48.2} & 46.0 & 40.2 & \bestcell{50.7} \\
OfficeQA & Acc & 45.3 & 49.7 & 47.5 & 53.1 & 54.6 & \secondcell{59.5} & 56.4 & 51.2 & \bestcell{62.8} \\
SpreadsheetBench & Acc & 44.3 & 46.1 & 45.8 & 49.6 & 52.5 & \secondcell{54.3} & 51.3 & 48.3 & \bestcell{56.8} \\
LiveMath & Acc & 35.2 & 36.8 & 37.4 & 41.5 & 50.8 & 56.0 & \secondcell{56.9} & 46.5 & \bestcell{61.3} \\
\midrule
\multicolumn{11}{@{}l}{\textit{DeepSeek-V4-Pro}} \\
\multirow{2}{*}{SkillsBench} & Soft Acc & 21.1 & 30.3 & 25.7 & 36.9 & 39.5 & \secondcell{44.2} & 43.9 & 34.3 & \bestcell{45.5} \\
 & Hard Acc & 13.1 & 19.9 & 16.5 & 26.5 & 28.4 & \secondcell{32.7} & 31.8 & 25.0 & \bestcell{34.2} \\
OfficeQA & Acc & 33.2 & 36.6 & 35.4 & 41.2 & 42.5 & \secondcell{46.8} & 44.3 & 40.7 & \bestcell{48.9} \\
SpreadsheetBench & Acc & 32.9 & 33.6 & 34.9 & 37.8 & 38.6 & \secondcell{41.2} & 39.7 & 38.1 & \bestcell{42.4} \\
LiveMath & Acc & 27.4 & 28.6 & 28.1 & 31.2 & 34.8 & \secondcell{37.5} & 36.2 & 32.0 & \bestcell{39.4} \\
\midrule
\multicolumn{11}{@{}l}{\textit{Qwen3-30B-A3B}} \\
SkillsBench & Soft Acc & 16.4 & 17.5 & 16.9 & 17.8 & 18.5 & \secondcell{19.8} & 19.2 & 17.9 & \bestcell{20.5} \\
OfficeQA & Acc & 25.8 & 27.6 & 26.1 & 28.0 & 28.9 & \secondcell{30.8} & 28.6 & 30.2 & \bestcell{32.5} \\
SpreadsheetBench & Acc & 30.0 & 30.0 & 30.0 & 30.6 & 31.0 & \secondcell{33.8} & 31.5 & 30.8 & \bestcell{36.4} \\
LiveMath & Acc & 23.6 & 24.1 & 23.9 & 25.8 & 27.5 & \secondcell{30.2} & 28.4 & 26.2 & \bestcell{33.1} \\
\bottomrule
\end{tabular}%
}

% \vspace{-0.15em}
\textbf{Claude Code harness}\\[-0.08em]
\resizebox{\linewidth}{!}{%
\begin{tabular}{@{}llccccccccc@{}}
\toprule
\multicolumn{2}{@{}l}{\textbf{Benchmark}} & \textbf{No skill} & \textbf{Human skill} & \textbf{LLM skill} & \textbf{GEPA} & \textbf{SkillEvolver} & \textbf{Memento-Skills} & \textbf{EvoSkill} & \textbf{SkillOpt} & \textbf{\method{}} \\
\midrule
\multicolumn{11}{@{}l}{\textit{Claude-Opus-4.6}} \\
\multirow{2}{*}{SkillsBench} & Soft Acc & 51.4 & 61.1 & 59.2 & 72.5 & 76.8 & \secondcell{81.2} & 79.5 & 80.2 & \bestcell{82.8} \\
 & Hard Acc & 40.0 & 46.6 & 43.9 & 52.9 & 56.5 & \secondcell{59.8} & 59.0 & 58.3 & \bestcell{60.9} \\
OfficeQA & Acc & 57.6 & 61.2 & 60.4 & 66.5 & 69.8 & \secondcell{74.8} & \bestcell{75.4} & 71.5 & 73.1 \\
SpreadsheetBench & Acc & 19.7 & 34.7 & 33.2 & 55.3 & 72.6 & 78.7 & 66.7 & \secondcell{80.0} & \bestcell{81.5} \\
LiveMath & Acc & 38.5 & 42.0 & 42.8 & 46.0 & 51.5 & \secondcell{54.4} & 50.2 & 53.7 & \bestcell{57.8} \\
\midrule
\multicolumn{11}{@{}l}{\textit{DeepSeek-V4-Pro}} \\
\multirow{2}{*}{SkillsBench} & Soft Acc & 28.8 & 34.0 & 32.6 & 42.6 & 46.3 & 50.1 & 49.0 & \secondcell{50.4} & \bestcell{51.6} \\
 & Hard Acc & 19.4 & 22.1 & 20.5 & 27.8 & 29.5 & 33.3 & 30.9 & \secondcell{33.8} & \bestcell{35.1} \\
OfficeQA & Acc & 38.9 & 43.1 & 41.5 & 44.0 & 46.2 & \secondcell{48.8} & 48.3 & 45.9 & \bestcell{50.0} \\
SpreadsheetBench & Acc & 24.7 & 30.8 & 31.2 & 38.5 & 41.7 & 44.9 & 42.8 & \secondcell{46.0} & \bestcell{46.9} \\
LiveMath & Acc & 29.5 & 32.0 & 31.5 & 35.2 & 38.6 & \bestcell{41.5} & 39.0 & \secondcell{40.8} & 39.4 \\
\bottomrule
\end{tabular}%
}
\vspace{-0.5em}
\end{table}

Table~\ref{tab:main_results} shows that \method{} consistently improves agent performance across different benchmarks and target models. Compared with no-skill baseline and existing skill optimization methods, \method{} achieves substantial gains in both no-harness and Claude Code settings, demonstrating the effectiveness of managing skills as a structured library rather than optimizing a single shared document. The improvements are also consistent across Claude-Opus-4.6, DeepSeek-V4-Pro, and Qwen3-30B-A3B, suggesting that the benefit of library-level optimization is not tied to a specific target model.
Meanwhile, Static human-written skills and one-shot LLM skills provide only limited improvements, while trajectory-driven methods perform better but remain below \method{}. This indicates that the main benefit comes from continuously managing which skills should be retrieved, updated, and retired. In particular, the advantage is more pronounced in heterogeneous settings such as SkillsBench, where task-specific feedback can conflict when compressed into a single skill document.

\subsection{Ablation Study}

\makeatletter
\newsavebox{\selmablbox}
\newsavebox{\selmevobox}
\newlength{\selmAblH}
\newlength{\selmEvoH}
\newlength{\selmAblW}
\newlength{\selmEvoW}
\setlength{\selmAblW}{0.40\textwidth}
\setlength{\selmEvoW}{0.58\textwidth}
\def\selmAblationRows{%
\begin{tabular}{@{}lc@{}}
\toprule
\textbf{Setting} & \textbf{Acc (\%)} \\
\midrule
Full (default) & \textbf{56.5} \\
\mbox{w/o} meta skill & 53.1 \\
\mbox{w/o} meta skill + slow update & 42.7 \\
\mbox{w/o} utility score & 46.3 \\
\mbox{w/o} utility-aware retrieval ($w_u{=}0$) & 49.2 \\
\mbox{w/o} validation gate & 42.5 \\
\mbox{w/o} graveyard & 50.4 \\
\bottomrule
\end{tabular}%
}
\def\selmEvolutionRows{%
\begin{tabular}{@{}lcccccc@{}}
\toprule
\textbf{Epoch} & \textbf{Active} & \textbf{ADD} & \textbf{EDIT} & \textbf{DOWNVOTE} & \textbf{Graveyard} & \textbf{Acc.\ (\%)} \\
\midrule
Init & 5 & -- & -- & -- & 0 & 17.5 \\
1 & 10 & 5 & 0 & 3 & 0 & 17.5 \\
2 & 11 & 2 & 0 & 4 & -- & 30.0 \\
Slow update & 11 & 2 & 2 & 2 & 3 & -- \\
\midrule
Final & 11 & 9 (total) & 2 (total) & 9 (total) & 3 & 36.43 \\
\bottomrule
\end{tabular}%
}
\setlength{\extrarowheight}{0pt}
\sbox{\selmablbox}{%
  \scriptsize\setlength{\tabcolsep}{2.4pt}\renewcommand{\arraystretch}{1}%
  \selmAblationRows
}
\sbox{\selmevobox}{%
  \scriptsize\setlength{\tabcolsep}{2pt}\renewcommand{\arraystretch}{1}%
  \selmEvolutionRows
}
\setlength{\selmAblH}{\dimexpr\ht\selmablbox+\dp\selmablbox\relax}
\setlength{\selmEvoH}{\dimexpr\ht\selmevobox+\dp\selmevobox\relax}
\ifdim\selmAblH>\selmEvoH
  % 6 = rows in the evolution tabular (header + 5 body rows)
  \setlength{\extrarowheight}{\dimexpr(\selmAblH-\selmEvoH)/6\relax}
  \sbox{\selmevobox}{%
    \scriptsize\setlength{\tabcolsep}{2pt}\renewcommand{\arraystretch}{1}%
    \selmEvolutionRows
  }%
\else
  % 8 = rows in the ablation tabular (header + 7 body rows)
  \setlength{\extrarowheight}{\dimexpr(\selmEvoH-\selmAblH)/8\relax}
  \sbox{\selmablbox}{%
    \scriptsize\setlength{\tabcolsep}{2.4pt}\renewcommand{\arraystretch}{1}%
    \selmAblationRows
  }%
\fi
\setlength{\extrarowheight}{0pt}
\Gscale@div\selmScaleA{\selmAblW}{\wd\selmablbox}
\Gscale@div\selmScaleB{\selmEvoW}{\wd\selmevobox}
\def\selmScale{1}
\ifdim\selmScaleA pt<\selmScale pt
  \let\selmScale\selmScaleA
\fi
\ifdim\selmScaleB pt<\selmScale pt
  \let\selmScale\selmScaleB
\fi
\ifdim\selmScale pt<1pt
  \sbox{\selmablbox}{\resizebox{\selmScale\wd\selmablbox}{!}{\usebox{\selmablbox}}}
  \sbox{\selmevobox}{\resizebox{\selmScale\wd\selmevobox}{!}{\usebox{\selmevobox}}}
\fi
\makeatother

\begin{table}[t]
\vspace{-0.5em}
\centering
\begin{minipage}[b]{\selmAblW}
  \centering
  \captionsetup{skip=3pt,font=small}
  \caption{Component ablation on SpreadsheetBench (Claude-Opus-4.6, no harness). \mbox{w/o}: without.}
  \label{tab:ablation}\par\nointerlineskip
  \usebox{\selmablbox}
\end{minipage}\hfill%
\begin{minipage}[b]{\selmEvoW}
  \centering
  \captionsetup{skip=3pt,font=small}
  \caption{Skill library evolution on SpreadsheetBench (Qwen3-30B-A3B, no harness). Init and epoch accuracies are on $D_{\mathrm{sel}}$; Final is held-out test.}
  \label{tab:evolution}\par\nointerlineskip
  \usebox{\selmevobox}
\end{minipage}
\vspace{-0.5em}
\end{table}

\begin{figure}[t]
\centering
\includegraphics[width=\linewidth]{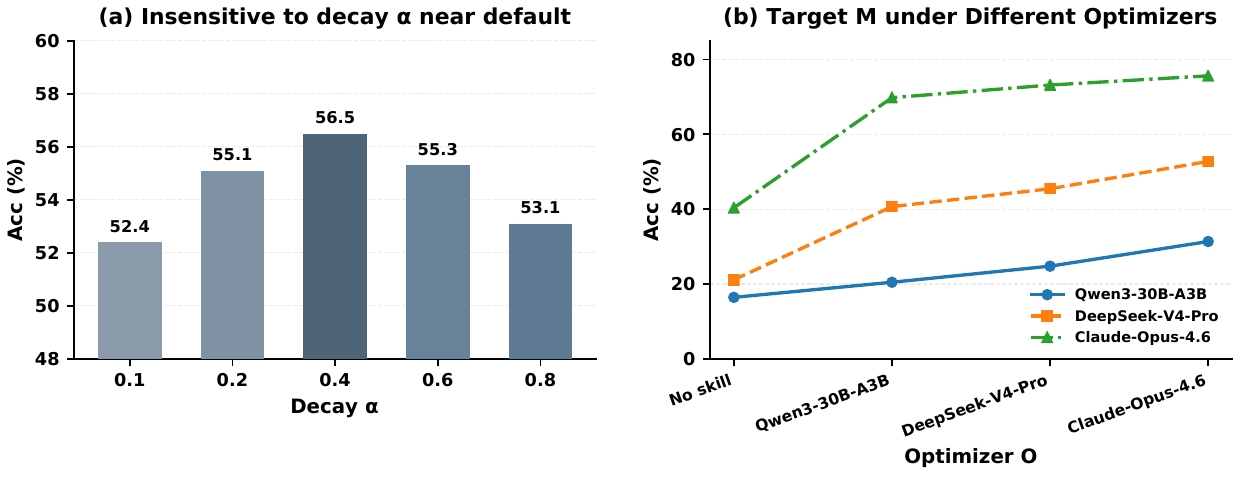}
\caption{(a)~Sensitivity to the EMA decay $\alpha$ on SpreadsheetBench (Claude-Opus-4.6, no harness); the marked default is $\alpha{=}0.4$. (b)~SkillsBench soft accuracy (\%) for each target model $M$ under different optimizers $O$, including the no-skill baseline (no harness).}
\label{fig:sensitivity}
\vspace{-0.5em}
\end{figure}

We ablate three aspects of \method{}: library-management components, utility estimation, and optimizer--target model interaction. We first remove individual management mechanisms, including validation, utility-aware retrieval, meta skills, slow updates, and the graveyard. We then vary the EMA decay factor $\alpha$ to examine the sensitivity of utility estimation. Finally, we vary the optimizer and target models to study how optimizer capability affects the resulting skill library. Unless otherwise specified, the experiments follow the default configuration.

\textbf{Library-management controls are critical to the effectiveness of skill evolution.} Table~\ref{tab:ablation} shows that removing the validation gate causes the largest single drop, while removing the utility score and utility-aware retrieval reduces only 2.9 points. Removing both meta skills and slow updates also leads to a large degradation, compared with 53.1\% when removing meta skills alone. The graveyard provides a smaller but consistent gain, with accuracy dropping to 50.4\% without it. These results show that validation, utility-aware selection, and longitudinal library updates are important for preventing ineffective skill changes from accumulating.

\textbf{\method{} is relatively insensitive to the EMA decay factor around the default setting. }As shown in Figure~\ref{fig:sensitivity}(a), $\alpha{=}0.4$ achieves 56.5\% accuracy, while $\alpha{=}0.2$ and $0.6$ obtain 55.1\% and 55.3\%, respectively. More extreme values, $\alpha{=}0.1$ and $0.8$, reduce accuracy to 52.4\% and 53.1\%. This indicates that the performance gain is not dependent on a narrowly tuned smoothing coefficient.

\textbf{A stronger optimizer generally produces a more effective skill library, while weaker optimizers can still benefit stronger target models.} Figure~\ref{fig:sensitivity}(b) shows that, with Qwen3-30B-A3B as the target, replacing the optimizer with DeepSeek-V4-Pro and Claude-Opus-4.6 increases soft accuracy from 20.45\% to 24.75\% and 31.34\%, respectively. The same trend holds for DeepSeek-V4-Pro as the target. Meanwhile, using Qwen3-30B-A3B as the optimizer still improves Claude-Opus-4.6 from the no-skill baseline.

\subsection{Library Evolution Analysis}

Table~\ref{tab:evolution} shows how the skill library changes during evolution on SpreadsheetBench with Qwen3-30B-A3B. The initial library contains 5 skills and achieves 17.5\% accuracy on the selection set. In Epoch 1, five new skills are added, but the accuracy remains unchanged. This shows that simply adding more skills does not improve performance.
In Epoch 2, only two new skills are added while four skills are downvoted, and the accuracy increases to 30.0\%. The slow update then makes two edits, adds two meta skills, and downvotes two more skills. Three skills are eventually moved to the graveyard because their confidence becomes non-positive, and they are therefore excluded from future retrieval.
The final library achieves 36.43\% accuracy, compared with 30.0\% for the no-skill baseline. These changes show that the library does not improve simply by growing in size; useful skills are added, while skills that repeatedly receive negative feedback are removed or revised.

% ============================================================
\section{Conclusion}
\label{sec:conclusion}
% ============================================================
We propose \method{}, a framework that manages skills as a library. 
%Through library management, Map-Reduce skill updates, and utility-aware retrieval, \method{} 
%Through library management that tracks confidence and utility, validates updates, and retires failed skills, \method{} prevents useful skills from being overwritten and ineffective skills from remaining in the library. 
\method{} tracks skill confidence and utility, validates skill updates, and retires failed skills.
Through the Map-Reduce framework and utility-aware retrieval, 
%\method{} reduces interference between skills for different types of tasks. Through utility-aware retrieval, 
\method{} helps the agent select the most relevant and effective skills.
%that are not only relevant but also effective in practice. 
Various experiments validate that \method{} consistently improves frozen agents over prior skill-optimization methods and achieves the strongest performance in most settings.

\bibliography{references}

@misc{yang2026skillopt,
  title={SkillOpt: Executive Strategy for Self-Evolving Agent Skills},
  author={Yifan Yang and Ziyang Gong and Weiquan Huang and Qihao Yang and Ziwei Zhou and Zisu Huang and Yan Li and Xuemei Gao and Qi Dai and Bei Liu and Kai Qiu and Yuqing Yang and Dongdong Chen and Xue Yang and Chong Luo},
  year={2026},
  eprint={2605.23904},
  archivePrefix={arXiv},
  primaryClass={cs.AI},
  url={https://arxiv.org/abs/2605.23904}
}

@misc{ni2026trace2skilldistilltrajectorylocallessons,
  title={Trace2Skill: Distill Trajectory-Local Lessons into Transferable Agent Skills},
  author={Jingwei Ni and Yihao Liu and Xinpeng Liu and Yutao Sun and Mengyu Zhou and Pengyu Cheng and Dexin Wang and Erchao Zhao and Xiaoxi Jiang and Guanjun Jiang},
  year={2026},
  eprint={2603.25158},
  archivePrefix={arXiv},
  primaryClass={cs.AI},
  url={https://arxiv.org/abs/2603.25158}
}

@misc{alzubi2026evoskill,
  title={EvoSkill: Automated Skill Discovery for Multi-Agent Systems},
  author={Salaheddin Alzubi and Noah Provenzano and Jaydon Bingham and Weiyuan Chen and Tu Vu},
  year={2026},
  eprint={2603.02766},
  archivePrefix={arXiv},
  primaryClass={cs.AI},
  url={https://arxiv.org/abs/2603.02766}
}

@misc{opsahlong2026officeqaproenterprisebenchmark,
  title={OfficeQA Pro: An Enterprise Benchmark for End-to-End Grounded Reasoning},
  author={Krista Opsahl-Ong and Arnav Singhvi and Jasmine Collins and Ivan Zhou and Cindy Wang and Ashutosh Baheti and Owen Oertell and Jacob Portes and Sam Havens and Erich Elsen and Michael Bendersky and Matei Zaharia and Xing Chen},
  year={2026},
  eprint={2603.08655},
  archivePrefix={arXiv},
  primaryClass={cs.AI},
  url={https://arxiv.org/abs/2603.08655}
}

@misc{zhang2026coevoskills,
  title={CoEvoSkills: Self-Evolving Agent Skills via Co-Evolutionary Verification},
  author={Hanrong Zhang and Shicheng Fan and Henry Peng Zou and Yankai Chen and Zhenting Wang and Jiayu Zhou and Chengze Li and Wei-Chieh Huang and Yifei Yao and Kening Zheng and Xue and Liu and Xiaoxiao Li and Philip S. Yu},
  year={2026},
  eprint={2604.01687},
  archivePrefix={arXiv},
  primaryClass={cs.AI},
  url={https://arxiv.org/abs/2604.01687}
}

@article{yuksekgonul2025textgrad,
  title={Optimizing generative {AI} by backpropagating language model feedback},
  author={Yuksekgonul, Mert and Bianchi, Federico and Boen, Joseph and Liu, Sheng and Lu, Pan and Huang, Zhi and Guestrin, Carlos and Zou, James},
  journal={Nature},
  volume={639},
  pages={609--616},
  year={2025},
  doi={10.1038/s41586-025-08661-4}
}

@inproceedings{agrawal2026gepa,
  author = {Agrawal, Lakshya A and Tan, Shangyin and Soylu, Dilara and Ziems, Noah and Khare, Rishi and Opsahl-Ong, Krista and Singhvi, Arnav and Shandilya, Herumb and Ryan, Michael J and Jiang, Meng and Potts, Christopher and Sen, Koushik and Dimakis, Alex and Stoica, Ion and Klein, Dan and Zaharia, Matei and Khattab, Omar},
 booktitle = {International Conference on Learning Representations},
 editor = {C. Vondrick and B. Hariharan and C. Raffel and L. Pinto and D. Yang and A. Faust},
 pages = {8479--8565},
 title = {GEPA: Reflective Prompt Evolution Can Outperform Reinforcement Learning},
 url = {https://proceedings.iclr.cc/paper_files/paper/2026/file/0e9e708b6f48e14fd0ac29e167413f76-Paper-Conference.pdf},
 volume = {2026},
 year = {2026}
}

@inproceedings{suzgun2026dynamiccheatsheet,
  title={Dynamic Cheatsheet: Test-Time Learning with Adaptive Memory},
  author={Suzgun, Mirac  and
      Yuksekgonul, Mert  and
      Bianchi, Federico  and
      Jurafsky, Dan  and
      Zou, James},
  booktitle={Proceedings of the 19th Conference of the {E}uropean Chapter of the {A}ssociation for {C}omputational {L}inguistics (Volume 1: Long Papers)},
  editor={Demberg, Vera  and
      Inui, Kentaro  and
      Marquez, Llu{\'i}s},
  pages={7080--7106},
  year={2026},
  address={Rabat, Morocco},
  publisher={Association for Computational Linguistics},
  doi={10.18653/v1/2026.eacl-long.333},
  url={https://aclanthology.org/2026.eacl-long.333/}
}

@misc{anthropic2025agentskills,
  title={Equipping Agents for the Real World with Agent Skills},
  author={Anthropic},
  year={2025},
  month={oct},
  url={https://www.anthropic.com/engineering/equipping-agents-for-the-real-world-with-agent-skills},
  note={Engineering blog, 16 October 2025}
}

@inproceedings{zhao2024expel,
  title={ExpeL: LLM Agents Are Experiential Learners},
  author={Zhao, Andrew and Huang, Daniel and Xu, Quentin and Lin, Matthieu and Liu, Yong-Jin and Huang, Gao},
  booktitle={Proceedings of the AAAI Conference on Artificial Intelligence},
  volume={38},
  number={17},
  pages={19632--19642},
  year={2024},
  doi={10.1609/aaai.v38i17.29936}
}

@inproceedings{ma2024spreadsheetbench,
   author = {Ma, Zeyao and Zhang, Bohan and Zhang, Jing and Yu, Jifan and Zhang, Xiaokang and Zhang, Xiaohan and Luo, Sijia and Wang, Xi and Tang, Jie},
 booktitle = {Advances in Neural Information Processing Systems},
 doi = {10.52202/079017-3007},
 editor = {A. Globerson and L. Mackey and D. Belgrave and A. Fan and U. Paquet and J. Tomczak and C. Zhang},
 pages = {94871--94908},
 publisher = {Curran Associates, Inc.},
 title = {SpreadsheetBench: Towards Challenging Real World Spreadsheet Manipulation},
 url = {https://proceedings.neurips.cc/paper_files/paper/2024/file/ac840df270ac537dd74530a15c332684-Paper-Datasets_and_Benchmarks_Track.pdf},
 volume = {37},
 year = {2024}
}

@misc{he2026livemathematicianbench,
  title={LiveMathematicianBench: A Live Benchmark for Research-Level Mathematical Reasoning with Proof Sketches},
  author={Linyang He and Qiyao Yu and Hanze Dong and Baohao Liao and Xinxing Xu and Micah Goldblum and Jiang Bian and Nima Mesgarani},
  year={2026},
  eprint={2604.01754},
  archivePrefix={arXiv},
  primaryClass={cs.CL},
  url={https://arxiv.org/abs/2604.01754}
}

@inproceedings{liu2026skillforge,
  author = {Liu, Xingyan and Luo, Xiyue and Li, Linyu and Huang, Ganghong and Liu, Jianfeng and Qiao, Honglin},
title = {SkillForge: Forging Domain-Specific, Self-Evolving Agent Skills in Cloud Technical Support},
year = {2026},
isbn = {9798400725999},
publisher = {Association for Computing Machinery},
address = {New York, NY, USA},
url = {https://doi.org/10.1145/3805712.3808466},
doi = {10.1145/3805712.3808466},
booktitle = {Proceedings of the 49th International ACM SIGIR Conference on Research and Development in Information Retrieval},
pages = {4763–4768},
numpages = {6},
location = {Australia},
series = {SIGIR '26}
}

@inproceedings{zhang2025ace,
  author = {Zhang, Qizheng and Hu, Changran and Upasani, Shubhangi and Ma, Boyuan and Hong, Fenglu and Kamanuru, Vamsidhar and Rainton, Jay and Wu, Chen and Ji, Mengmeng and Li, Hanchen and Thakker, Urmish and Zou, James Y and Olukotun, Kunle},
 booktitle = {International Conference on Learning Representations},
 editor = {C. Vondrick and B. Hariharan and C. Raffel and L. Pinto and D. Yang and A. Faust},
 pages = {86069--86100},
 title = {Agentic Context Engineering: Evolving Contexts for Self-Improving Language Models},
 url = {https://proceedings.iclr.cc/paper_files/paper/2026/file/8a94ff6f922d995d7d3f4ebf4143e442-Paper-Conference.pdf},
 volume = {2026},
 year = {2026}
}

@misc{li2026skillsbench,
  title={SkillsBench: Benchmarking How Well Agent Skills Work Across Diverse Tasks}, 
      author={Xiangyi Li and Yimin Liu and Wenbo Chen and Bingran You and Zonglin Di and Yifeng He and Shenghan Zheng and Kyoung Whan Choe and Jiankai Sun and Shuyi Wang and Chujun Tao and Binxu Li and Xuandong Zhao and Hejia Geng and Xiaojun Wu and Junwei Zhou and Xiaokun Chen and Hanwen Xing and Yubo Li and Qunhong Zeng and Di Wang and Yuanli Wang and Roey Ben Chaim and Penghao Jiang and Haotian Shen and Luyang Kong and Xinyi Liu and Runhui Wang and Xuanqing Liu and Jiachen Li and Xin Lan and Yueqian Lin and Wengao Ye and Junwei He and Songlin Li and Yue Zhang and Yipeng Gao and Yijiang Li and Ze Ma and Liqiang Jing and Tianyu Wang and Kaixin Li and Yiqi Xue and Haoran Lyu and Yizhuo He and Yuchen Tian and Shutong Wu and Bowei Wang and Yixuan Gao and Bo Chen and Litong Liu and Sikai Cheng and Jiajun Bao and Shuaicheng Tong and Shuwen Xu and Terry Yue Zhuo and Tinghan Ye and Qi Qi and Miao Li and Longtai Liao and Zelin Tan and Chang Shi and Xilin Tang and Srinath Tankasala and Boqin Yuan and Yaoyao Qian and Jianhong Tu and Chenguang Wang and Yizhou Sun and Wei Wang and Aaron Taylor and Ziyue Yang and Changkun Guan and Zhikang Dong and Xinyu Zhang and Steven Dillmann and Han-chung Lee and Dawn Song},
      year={2026},
      eprint={2602.12670},
      archivePrefix={arXiv},
      primaryClass={cs.AI},
      url={https://arxiv.org/abs/2602.12670}
}

@misc{wang2023voyager,
  title={Voyager: An Open-Ended Embodied Agent with Large Language Models}, 
      author={Guanzhi Wang and Yuqi Xie and Yunfan Jiang and Ajay Mandlekar and Chaowei Xiao and Yuke Zhu and Linxi Fan and Anima Anandkumar},
      year={2023},
      eprint={2305.16291},
      archivePrefix={arXiv},
      primaryClass={cs.AI},
      url={https://arxiv.org/abs/2305.16291},
}

@misc{zheng2025skillweaver,
  title={SkillWeaver: Web Agents can Self-Improve by Discovering and Honing Skills}, 
      author={Boyuan Zheng and Michael Y. Fatemi and Xiaolong Jin and Zora Zhiruo Wang and Apurva Gandhi and Yueqi Song and Yu Gu and Jayanth Srinivasa and Gaowen Liu and Graham Neubig and Yu Su},
      year={2025},
      eprint={2504.07079},
      archivePrefix={arXiv},
      primaryClass={cs.AI},
      url={https://arxiv.org/abs/2504.07079}
}

@misc{ma2026skillgen,
  title={SkillGen: Verified Inference-Time Agent Skill Synthesis}, 
      author={Yuchen Ma and Yue Huang and Han Bao and Haomin Zhuang and Swadheen Shukla and Michel Galley and Xiangliang Zhang and Stefan Feuerriegel},
      year={2026},
      eprint={2605.10999},
      archivePrefix={arXiv},
      primaryClass={cs.LG},
      url={https://arxiv.org/abs/2605.10999}
}

@misc{zhang2026skillevolver,
  title={SkillEvolver: Skill Learning as a Meta-Skill}, 
      author={Genrui Zhang and Erle Zhu and Jinfeng Zhou and Caiyan Jia and Hongning Wang},
      year={2026},
      eprint={2605.10500},
      archivePrefix={arXiv},
      primaryClass={cs.AI},
      url={https://arxiv.org/abs/2605.10500}
}

@misc{zhou2026mementoskills,
  title={Memento-Skills: Let Agents Design Agents}, 
      author={Huichi Zhou and Siyuan Guo and Anjie Liu and Zhongwei Yu and Ziqin Gong and Bowen Zhao and Zhixun Chen and Menglong Zhang and Yihang Chen and Jinsong Li and Runyu Yang and Qiangbin Liu and Xinlei Yu and Jianmin Zhou and Na Wang and Chunyang Sun and Jun Wang},
      year={2026},
      eprint={2603.18743},
      archivePrefix={arXiv},
      primaryClass={cs.AI},
      url={https://arxiv.org/abs/2603.18743}
}

@misc{yang2026autoskill,
  title={AutoSkill: Experience-Driven Lifelong Learning via Skill Self-Evolution}, 
      author={Yutao Yang and Junsong Li and Qianjun Pan and Bihao Zhan and Yuxuan Cai and Lin Du and Jie Zhou and Kai Chen and Qin Chen and Xin Li and Bo Zhang and Liang He},
      year={2026},
      eprint={2603.01145},
      archivePrefix={arXiv},
      primaryClass={cs.AI},
      url={https://arxiv.org/abs/2603.01145}
}

@misc{xia2026skillrl,
  title={SkillRL: Evolving Agents via Recursive Skill-Augmented Reinforcement Learning}, 
      author={Peng Xia and Jianwen Chen and Hanyang Wang and Jiaqi Liu and Kaide Zeng and Yu Wang and Siwei Han and Yiyang Zhou and Xujiang Zhao and Haifeng Chen and Zeyu Zheng and Cihang Xie and Huaxiu Yao},
      year={2026},
      eprint={2602.08234},
      archivePrefix={arXiv},
      primaryClass={cs.LG},
      url={https://arxiv.org/abs/2602.08234}
}

@misc{jiang2026xskill,
  title={XSkill: Continual Learning from Experience and Skills in Multimodal Agents}, 
      author={Guanyu Jiang and Zhaochen Su and Xiaoye Qu and Yi R. Fung},
      year={2026},
      eprint={2603.12056},
      archivePrefix={arXiv},
      primaryClass={cs.AI},
      url={https://arxiv.org/abs/2603.12056}
}

@inproceedings{wu2026evolver,
  title={EvolveR: Self-Evolving LLM Agents through an Experience-Driven Lifecycle}, 
      author={Rong Wu and Xiaoman Wang and Jianbiao Mei and Pinlong Cai and Daocheng Fu and Cheng Yang and Licheng Wen and Xuemeng Yang and Yufan Shen and Yuxin Wang and Botian Shi},
      year={2026},
      eprint={2510.16079},
      archivePrefix={arXiv},
      primaryClass={cs.CL},
      url={https://arxiv.org/abs/2510.16079}
}

@misc{wang2026skillgrad,
  title={SkillGrad: Optimizing Agent Skills Like Gradient Descent}, 
      author={Hanyu Wang and Yifan Lan and Bochuan Cao and Lu Lin and Jinghui Chen},
      year={2026},
      eprint={2605.27760},
      archivePrefix={arXiv},
      primaryClass={cs.AI},
      url={https://arxiv.org/abs/2605.27760}
}

@misc{qiu2026autorefine,
  title={AutoRefine: Compiling Trajectories into Validated Typed Agent Artifacts}, 
      author={Libin Qiu and Zhirong Gao and Junfu Chen and Yuhang Ye and Liangyu Li and Weizhi Huang and Xiaobo Xue and Wenkai Qiu and Shuo Tang},
      year={2026},
      eprint={2601.22758},
      archivePrefix={arXiv},
      primaryClass={cs.AI},
      url={https://arxiv.org/abs/2601.22758}
}

@misc{pu2026skillops,
  title={SkillOps: Managing LLM Agent Skill Libraries as Self-Maintaining Software Ecosystems}, 
      author={Hongji Pu and Xinyuan Song and Liang Zhao},
      year={2026},
      eprint={2605.13716},
      archivePrefix={arXiv},
      primaryClass={cs.SE},
      url={https://arxiv.org/abs/2605.13716}
}

@misc{lin2026museautoskill,
  title={MUSE-Autoskill: Self-Evolving Agents via Skill Creation, Memory, Management, and Evaluation}, 
      author={Huawei Lin and Peng Li and Jie Song and Fuxin Jiang and Tieying Zhang},
      year={2026},
      eprint={2605.27366},
      archivePrefix={arXiv},
      primaryClass={cs.AI},
      url={https://arxiv.org/abs/2605.27366}
}

@misc{zhang2026skillflow,
  title={SkillFlow:Benchmarking Lifelong Skill Discovery and Evolution for Autonomous Agents}, 
      author={Ziao Zhang and Kou Shi and Shiting Huang and Avery Nie and Yu Zeng and Yiming Zhao and Zhen Fang and Qishen Su and Haibo Qiu and Wei Yang and Qingnan Ren and Shun Zou and Wenxuan Huang and Lin Chen and Zehui Chen and Feng Zhao},
      year={2026},
      eprint={2604.17308},
      archivePrefix={arXiv},
      primaryClass={cs.AI},
      url={https://arxiv.org/abs/2604.17308}
}

@misc{anthropic2025claudecode,
  title={{Claude Code}},
  author={{Anthropic}},
  year={2025},
  url={https://code.claude.com/docs/en/overview},
  note={Official documentation}
}

@misc{anthropic2026claudeopus46,
  title={Introducing Claude Opus 4.6},
  author={Anthropic},
  year={2026},
  month={feb},
  url={https://www.anthropic.com/research/claude-opus-4-6},
}

@misc{deepseekai2026deepseekv4,
  title={DeepSeek-V4: Towards Highly Efficient Million-Token Context Intelligence}, 
      author={DeepSeek-AI},
      year={2026},
      eprint={2606.19348},
      archivePrefix={arXiv},
      primaryClass={cs.CL},
      url={https://arxiv.org/abs/2606.19348}
}

@misc{qwen2025qwen3,
  title={Qwen3 Technical Report}, 
      author={An Yang and Anfeng Li and Baosong Yang and Beichen Zhang and Binyuan Hui and Bo Zheng and Bowen Yu and Chang Gao and Chengen Huang and Chenxu Lv and Chujie Zheng and Dayiheng Liu and Fan Zhou and Fei Huang and Feng Hu and Hao Ge and Haoran Wei and Huan Lin and Jialong Tang and Jian Yang and Jianhong Tu and Jianwei Zhang and Jianxin Yang and Jiaxi Yang and Jing Zhou and Jingren Zhou and Junyang Lin and Kai Dang and Keqin Bao and Kexin Yang and Le Yu and Lianghao Deng and Mei Li and Mingfeng Xue and Mingze Li and Pei Zhang and Peng Wang and Qin Zhu and Rui Men and Ruize Gao and Shixuan Liu and Shuang Luo and Tianhao Li and Tianyi Tang and Wenbiao Yin and Xingzhang Ren and Xinyu Wang and Xinyu Zhang and Xuancheng Ren and Yang Fan and Yang Su and Yichang Zhang and Yinger Zhang and Yu Wan and Yuqiong Liu and Zekun Wang and Zeyu Cui and Zhenru Zhang and Zhipeng Zhou and Zihan Qiu},
      year={2025},
      eprint={2505.09388},
      archivePrefix={arXiv},
      primaryClass={cs.CL},
      url={https://arxiv.org/abs/2505.09388}
}
\bibliographystyle{iclr2027_conference}

% ============================================================
\appendix

\section{Iterative Library Evolution Algorithm}
\label{app:algorithm}

Algorithm~\ref{alg:main} summarizes the \method{} loop in \S\ref{sec:method}: utility-aware retrieval and rollout, diagnostic trajectory selection, Map--Reduce CRUD synthesis, validation gating, pruning to the graveyard, and slow updates.

\begin{algorithm}[t]
\caption{\textsc{Selm} --- Iterative Library Evolution}
\label{alg:main}
\small
\begin{algorithmic}[1]
\REQUIRE Frozen target model $M$, execution harness $h$,
training split $D_{\mathrm{tr}}$, selection split $D_{\mathrm{sel}}$,
epochs $E$, retrieval budget $K$, trajectory budget $B$,
EMA coefficient $\alpha$, utility weight $w_u$,
retirement threshold $\tau_u$, maximum capacity $M_{\max}$
\STATE $L_0 \gets$ initial skill library \hfill $\triangleright$ empty or seed skills
\STATE $G_0 \gets \emptyset$ \hfill $\triangleright$ graveyard starts empty
\STATE $L_{\mathrm{best}} \gets L_0$ \hfill $\triangleright$ snapshot kept for export
\STATE $V_{\mathrm{best}} \gets V(L_0)$ \hfill $\triangleright$ score on $D_{\mathrm{sel}}$
\FOR{$e=1,\ldots,E$}
    \FOR{each training batch $\mathcal{B}\subset D_{\mathrm{tr}}$ \hfill $\triangleright$ one fast update}
        \STATE $S_t \gets \{s\in L_t : n(s)\notin\{n(g)\mid g\in G_t\}\}$ \hfill $\triangleright$ block retired names
        \FOR{each task $x_i\in\mathcal{B}$}
            \STATE Let $\gamma_i$ be the category of $x_i$
            \STATE $S_t(x_i) \gets \operatorname{TopK}\bigl(\{s\in S_t : \gamma(s)\in\{\gamma_i,\mathrm{general}\}\}, R(s,x_i), K\bigr)$ \hfill $\triangleright$ utility-aware retrieval
            \STATE $(\tau_i, r_{\mathrm{hard},i}, r_{\mathrm{soft},i}) \gets h(M, x_i, S_t(x_i))$ \hfill $\triangleright$ rollout
            \STATE $r_i \gets \mathcal{G}(r_{\mathrm{hard},i}, r_{\mathrm{soft},i})$ \hfill $\triangleright$ scalar reward
            \STATE $p_i \gets \bigl|r_i-\bar{u}(S_t(x_i))\bigr|$ \hfill $\triangleright$ gap from stored utility
        \ENDFOR
        \STATE Select the top-$B$ trajectories according to $p_i$ \hfill $\triangleright$ most informative traces
        \FOR{each skill $s$ used in the current batch}
            \STATE $u(s)\gets (1-\alpha)u(s)+\alpha\bar{r}_s$ \hfill $\triangleright$ EMA utility
        \ENDFOR
        \STATE Partition selected trajectories by task category \hfill $\triangleright$ Map
        \FOR{each category $c$}
            \STATE $a_c \gets \operatorname{TrajectoryAnalyst}(\text{trajectories of }c, L_t)$ \hfill $\triangleright$ success, failure, insights
        \ENDFOR
        \STATE $\mathrm{ops}_t \gets \operatorname{SkillSynthesizer}(\{a_c\}, L_t)$ \hfill $\triangleright$ Reduce; operation budgets
        \STATE $L'_t \gets \operatorname{Apply}(L_t, \mathrm{ops}_t, G_t)$ \hfill $\triangleright$ candidate library
        \STATE Evaluate $V(L'_t)=\frac{1}{|D_{\mathrm{sel}}|}\sum_{x\in D_{\mathrm{sel}}} r(L'_t, x)$ \hfill $\triangleright$ validation gate
        \IF{$V(L'_t)>V(L_t)$}
            \STATE $L_t\gets L'_t$ \hfill $\triangleright$ accept only if improved
        \ENDIF
        \IF{$V(L_t)>V_{\mathrm{best}}$}
            \STATE $L_{\mathrm{best}}\gets L_t$ \hfill $\triangleright$ keep the best snapshot
            \STATE $V_{\mathrm{best}}\gets V(L_t)$
        \ENDIF
    \ENDFOR
    \IF{slow update is scheduled at epoch $e$ \hfill $\triangleright$ less frequent than fast updates}
        \STATE Generate additional CRUD operations \hfill $\triangleright$ consolidate persistent trends
        \STATE $L'_t\gets \operatorname{Apply}(L_t, \mathrm{ops}^{\mathrm{slow}}_t, G_t)$
        \STATE Validate $L'_t$ on $D_{\mathrm{sel}}$ \hfill $\triangleright$ same gate as the fast update
        \IF{$V(L'_t)>V(L_t)$}
            \STATE $L_t\gets L'_t$
        \ENDIF
        \IF{$V(L_t)>V_{\mathrm{best}}$}
            \STATE $L_{\mathrm{best}}\gets L_t$
            \STATE $V_{\mathrm{best}}\gets V(L_t)$
        \ENDIF
    \ENDIF
    \STATE \textbf{Lifecycle pruning:}
    \FOR{each $s\in L_t$}
        \IF{$u(s)<\tau_u$ \textbf{or} $v(s)\le 0$ \hfill $\triangleright$ low utility or non-positive confidence}
            \STATE $L_t\gets L_t\setminus\{s\}$
            \STATE $G_t\gets G_t\cup\{s\}$ \hfill $\triangleright$ retire; name cannot return
        \ENDIF
    \ENDFOR
    \STATE Enforce $|G_t|\le |G|_{\max}$ \hfill $\triangleright$ graveyard capacity
    \STATE Enforce $|S_t|\le M_{\max}$, name uniqueness, and graveyard blocking
\ENDFOR
\RETURN $L_{\mathrm{best}}$ \hfill $\triangleright$ not necessarily the last $L_t$
\end{algorithmic}
\end{algorithm}

\section{Benchmark Details}
\label{app:benchmarks}

We evaluate on four agent benchmarks covering heterogeneous domains, procedural tool use, long-context document reasoning, and mathematical problem solving.

\textbf{SkillsBench}~\citep{li2026skillsbench}: a multi-domain benchmark with 8 task categories (cybersecurity, data analysis, finance-economics, hardware engineering, natural science, social science, software engineering, and one general category). SkillsBench reports both hard accuracy  and soft accuracy, corresponding to $r_{\mathrm{hard}}$ and $r_{\mathrm{soft}}$ in \S\ref{sec:retrieval}. 

\textbf{SpreadsheetBench}~\citep{ma2024spreadsheetbench}: spreadsheet manipulation requiring multi-step formula generation and cell-level operations.

\textbf{OfficeQA}~\citep{opsahlong2026officeqaproenterprisebenchmark}: a grounded reasoning benchmark built from U.S.\ Treasury Bulletins with a corpus of approximately 89,000 pages. 

\textbf{LiveMath}~\citep{he2026livemathematicianbench}: LiveMathematicianBench(LiveMath), a live mathematician-level reasoning benchmark, abbreviated LiveMath in tables. 

The last three benchmarks report a single accuracy.

\section{Hyperparameter Summary}
\label{app:hyperparams}

\begin{table}[h]
\caption{Complete hyperparameter summary with default values.}
\label{tab:hyperparams}
\begin{center}
\small
\begin{tabular}{llc}
\toprule
\textbf{Hyperparameter} & \textbf{Description} & \textbf{Default} \\
\midrule
$M_{\max}$ & Maximum library capacity & 30 \\
$w_u$ & Utility weight in retrieval scoring & 0.6 \\
$w$ & soft score weight in reward calculating & 0.5 \\
$\alpha$ & EMA decay factor for utility update & 0.4 \\
$\tau_\text{soft}$ & Soft threshold for success classification & 0.3 \\
$|\graveyard|_{\max}$ & Maximum graveyard capacity & 50 \\
$B_\text{ADD}$ & Max ADD operations per evolution step & 5 \\
$B_\text{EDIT}$ & Max EDIT operations per evolution step & 3 \\
$K$ & Number of skills retrieved per task & 3 \\
$B$ & Trajectory selection budget & 50 \\
$N_{slow}$ & Slow update sample size & 20 \\
$\tau_u$ & threshold for utility score & 0.05 \\
Injection budget & Max characters for skill injection & 2000 \\
Fail compression & Max chars for failed trajectories & 2000 \\
Success compression & Max chars for successful trajectories & 400 \\
Map parallelism & Max parallel TrajectoryAnalyst workers & 16 \\
\bottomrule
\end{tabular}
\end{center}
\end{table}

\section{Skill Library Example}
\label{app:skill_example}

Figures~\ref{fig:skill_output}--\ref{fig:skill_slow} reproduce four skills from the optimized SpreadsheetBench library (Qwen3-30B-A3B, no harness; 11 active skills, test accuracy 36.43\%), one from each remaining category. Text is the full library snapshot, not a truncation.

\begin{skillbox}{Skill: \texttt{output\_requirements}}
\small
\textbf{Category.} \texttt{general} \quad
\textbf{Confidence.} 4 \quad
\textbf{Utility.} 0.43

\textbf{Description.} Output Requirements.

\textbf{Trigger conditions.} None; injected whenever the skill is retrieved.

\textbf{Content.}
\begin{itemize}\setlength{\itemsep}{2pt}\setlength{\parskip}{0pt}\setlength{\topsep}{2pt}
    \item Save the result to \texttt{OUTPUT\_PATH}.
    \item Do not hardcode row counts or column letters --- iterate over actual rows in the workbook.
    \item Preserve sheets and cells not mentioned in the instruction.
\end{itemize}
\end{skillbox}
\captionof{figure}{A general skill from the optimized SpreadsheetBench library.}
\label{fig:skill_output}

\begin{skillbox}{Skill: \texttt{dynamic\_date\_handling}}
\small
\textbf{Category.} \texttt{cell\_level} \quad
\textbf{Confidence.} 2 \quad
\textbf{Utility.} 0.44

\textbf{Description.} Handle dynamic date ranges and exclude weekends/holidays in formulas.

\textbf{Trigger conditions.} When tasks involve date-range calculations with dynamic inputs or require exclusion of non-business days.

\textbf{Content.}
\begin{enumerate}\setlength{\itemsep}{2pt}\setlength{\parskip}{0pt}\setlength{\topsep}{2pt}
    \item Use \texttt{DATE} functions to define start and end dates.
    \item Use \texttt{NETWORKDAYS} or custom logic to exclude weekends.
    \item Include holiday lists if provided.
    \item Use \texttt{IF} statements to validate date ranges.
    \item Test with sample data to ensure accuracy.
\end{enumerate}
\end{skillbox}
\captionof{figure}{A cell-level skill from the optimized SpreadsheetBench library.}
\label{fig:skill_date}

\begin{skillbox}{Skill: \texttt{inter\_sheet\_data\_matching}}
\small
\textbf{Category.} \texttt{sheet\_level} \quad
\textbf{Confidence.} 2 \quad
\textbf{Utility.} 0.44

\textbf{Description.} Handle data matching and manipulation across multiple sheets.

\textbf{Trigger conditions.} When tasks require data from multiple sheets to be compared, combined, or processed.

\textbf{Content.}
\begin{enumerate}\setlength{\itemsep}{2pt}\setlength{\parskip}{0pt}\setlength{\topsep}{2pt}
    \item Use VBA to reference data from other sheets.
    \item Use loops and conditionals to match data.
    \item Use error handling to prevent runtime issues.
    \item Test with different data structures.
    \item Use message boxes to provide feedback on results.
\end{enumerate}
\end{skillbox}
\captionof{figure}{A sheet-level skill from the optimized SpreadsheetBench library.}
\label{fig:skill_sheet}

\begin{skillbox}{Skill: \texttt{slow\_update\_epoch\_02}}
\small
\textbf{Category.} \texttt{meta} \quad
\textbf{Confidence.} 3 \quad
\textbf{Utility.} 0.70

\textbf{Description.} Epoch-level guidance from epoch~2 slow update.

\textbf{Trigger conditions.} None; produced by the epoch-level slow update and managed by the same utility and lifecycle rules as ordinary skills.

\textbf{Content.}
\begin{itemize}\setlength{\itemsep}{3pt}\setlength{\parskip}{0pt}\setlength{\topsep}{2pt}
    \item \textbf{ADD}: \texttt{meta\_insight\_epoch\_2} --- Identified patterns of persistent failures and regressions, highlighting the need for better handling of dynamic data, text parsing, and formula execution in VBA.\\
    Content: Focus on improving dynamic data handling, text parsing, and formula execution. Address issues with date handling, string matching, and VBA macro behavior. Prioritize skills that were downvoted or showed persistent failure.
    \item \textbf{EDIT}: \texttt{dynamic\_column\_handling}
    \item \textbf{EDIT}: \texttt{advanced\_text\_parsing\_in\_vba}
    \item \textbf{DOWNVOTE}: \texttt{text\_matching\_strategies} (Showed persistent failure in text comparison and filtering tasks, indicating a need for re-evaluation of its implementation and use cases.)
    \item \textbf{DOWNVOTE}: \texttt{vba\_data\_preservation} (Caused regression in data preservation tasks, suggesting potential issues with how it handles macro execution and data integrity.)
\end{itemize}
\end{skillbox}

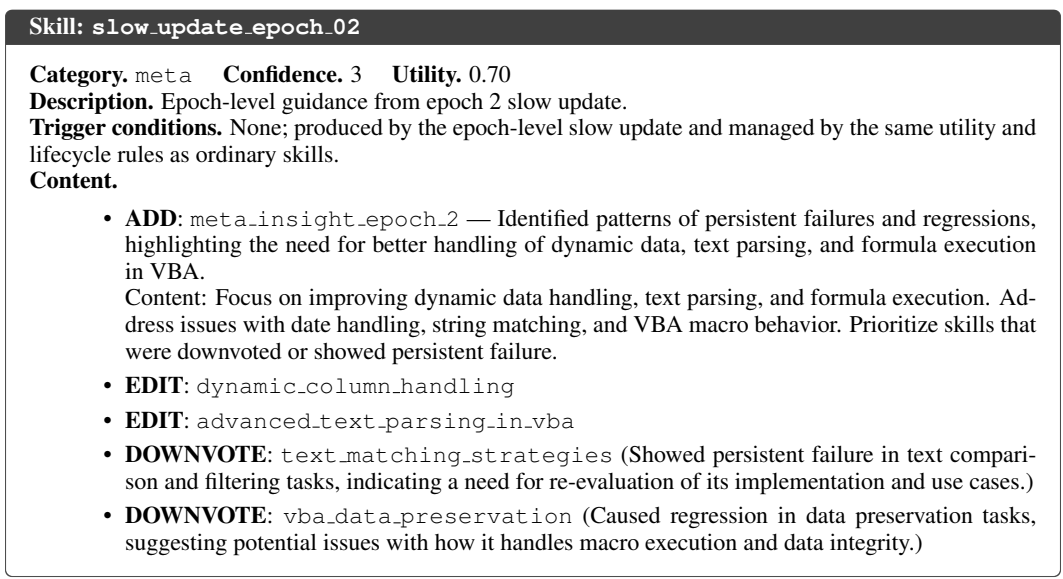
\captionof{figure}{A meta skill produced by the epoch-level slow update.}
\label{fig:skill_slow}

% ============================================================
\section{Limitations and Future Work}
\label{sec:limitations}
% ============================================================

\textbf{Semantic deduplication.} The graveyard mechanism operates at the name level (exact string match), preventing exact-name oscillation but not semantically equivalent skills with different names from coexisting. Embedding-based similarity detection would be a natural extension.

\textbf{Cross-environment transfer.} A skill library optimized on one benchmark could potentially warm-start optimization on another, analogous to transfer learning in neural networks. We leave exploration of cross-benchmark transfer to future work.

\textbf{Scalability.} While we demonstrate effectiveness with up to 30 skills across 8 categories, scaling to hundreds of categories or thousands of skills would require more sophisticated indexing (e.g., embedding-based retrieval) and potentially hierarchical library organization.

\textbf{Optimizer model dependence.} The quality of evolution depends on the optimizer LLM's ability to analyze trajectories and synthesize CRUD operations. Weaker optimizer models may produce lower-quality operations, and the interaction between optimizer capability and library quality deserves systematic study.

\end{document}